\documentclass[10pt,twocolumn,letterpaper]{article}

\usepackage{cvpr}
\usepackage{cvpr}
\usepackage{times}
\usepackage{epsfig}
\usepackage{graphicx}
\usepackage{amsmath}
\usepackage{amssymb}
\input{psfig.sty}

\newcommand{\MYF}{./}
\newcommand{\1}{{\bf 1}}

\usepackage[pagebackref=true,breaklinks=true,letterpaper=true,colorlinks,bookmarks=false]{hyperref}

\cvprfinalcopy 

\def\cvprPaperID{792} 

\ifcvprfinal\fi
\begin{document}

\title{Layered Logic Classifiers: Exploring the `And' and `Or' Relations}

\author{Zhuowen Tu$^1$, Piotr Dollar$^2$, and Yingnian Wu$^3$\\
$^1$Dept. of Cognitive Science, UCSD, $^2$Microsoft Research, $^3$Department of Statistics, UCLA\\
{\tt\small ztu@ucsd.edu,pdollar@gmail.com,ywu@stat.ucla.edu}
}

\maketitle

\begin{abstract}
Designing effective and efficient classifier for pattern analysis
is a key problem in machine learning and computer vision.
Many the solutions to the problem require to perform logic operations
such as `and', `or', and `not'. Classification and regression tree (CART)
include these operations explicitly.
Other methods such as neural networks, SVM, and boosting
learn/compute a weighted sum on features (weak classifiers), which
weakly perform the 'and' and 'or' operations.
However, it is hard for these classifiers to deal with the 'xor' pattern directly.
In this paper, we propose layered logic classifiers for patterns
of complicated distributions by combining the `and', `or', and `not' operations.
The proposed algorithm is very general and easy to implement.
We test the classifiers on several typical datasets from the
Irvine repository and two challenging vision applications, object segmentation
and pedestrian detection.
We observe significant improvements on all the datasets
over the widely used decision stump based AdaBoost algorithm. 
The resulting classifiers have much less training complexity than decision tree based AdaBoost,
and can be applied in a wide range of domains.

\end{abstract}

\section{Introduction}

\begin{figure}[!htbp]
\begin{center}
\begin{tabular}{cccc}
\hspace{-8mm} \psfig{figure=\MYF/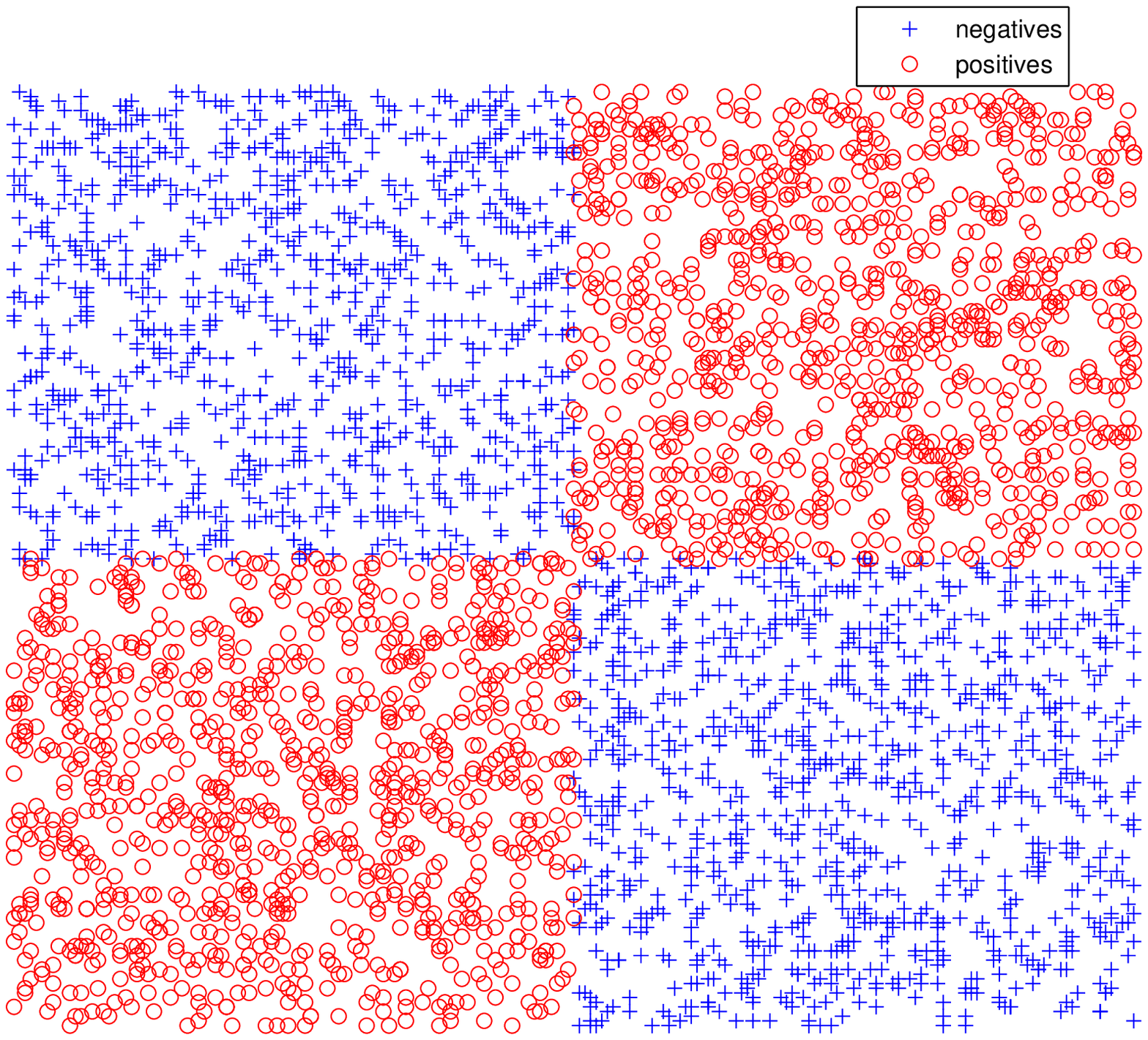,width=1.0in} &
\hspace{-8mm} \psfig{figure=\MYF/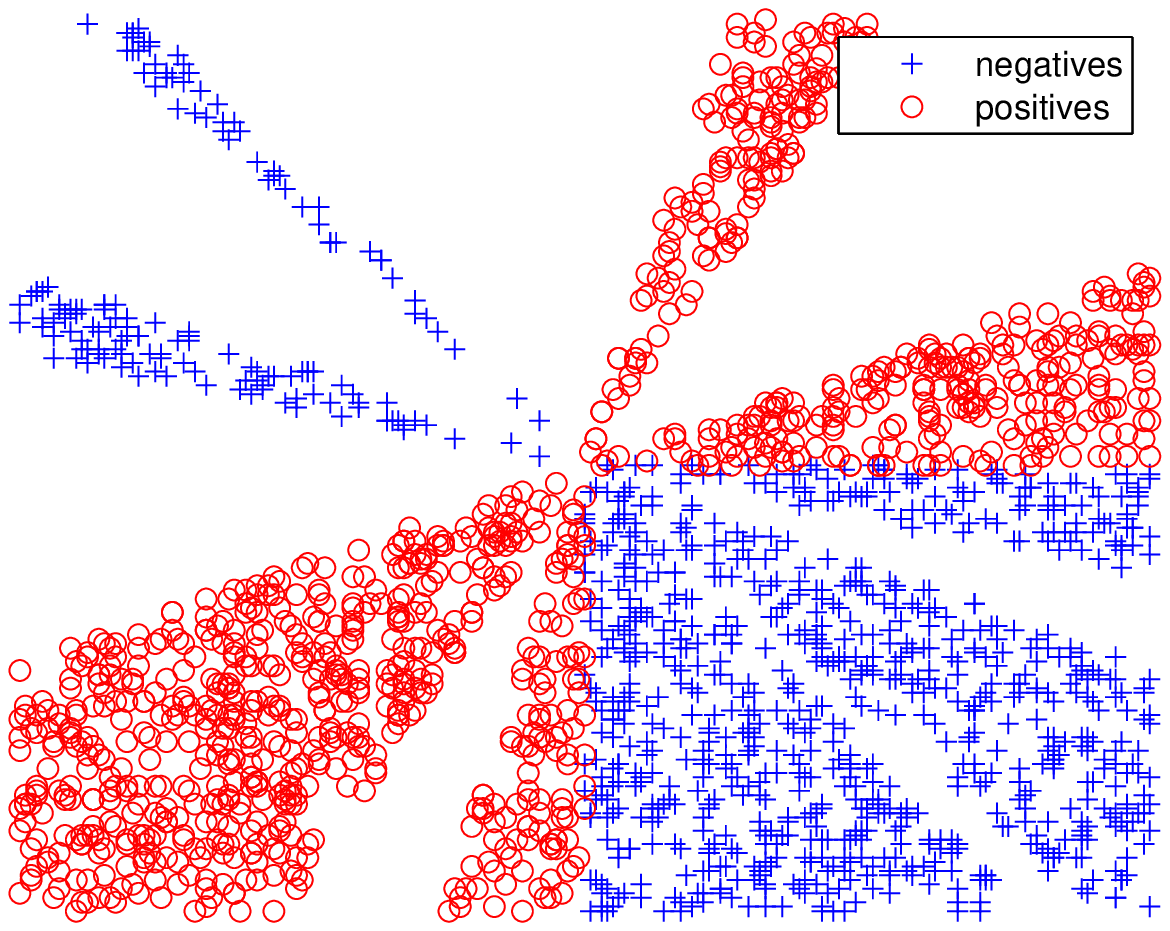,width=1.0in} &
\hspace{-8mm} \psfig{figure=\MYF/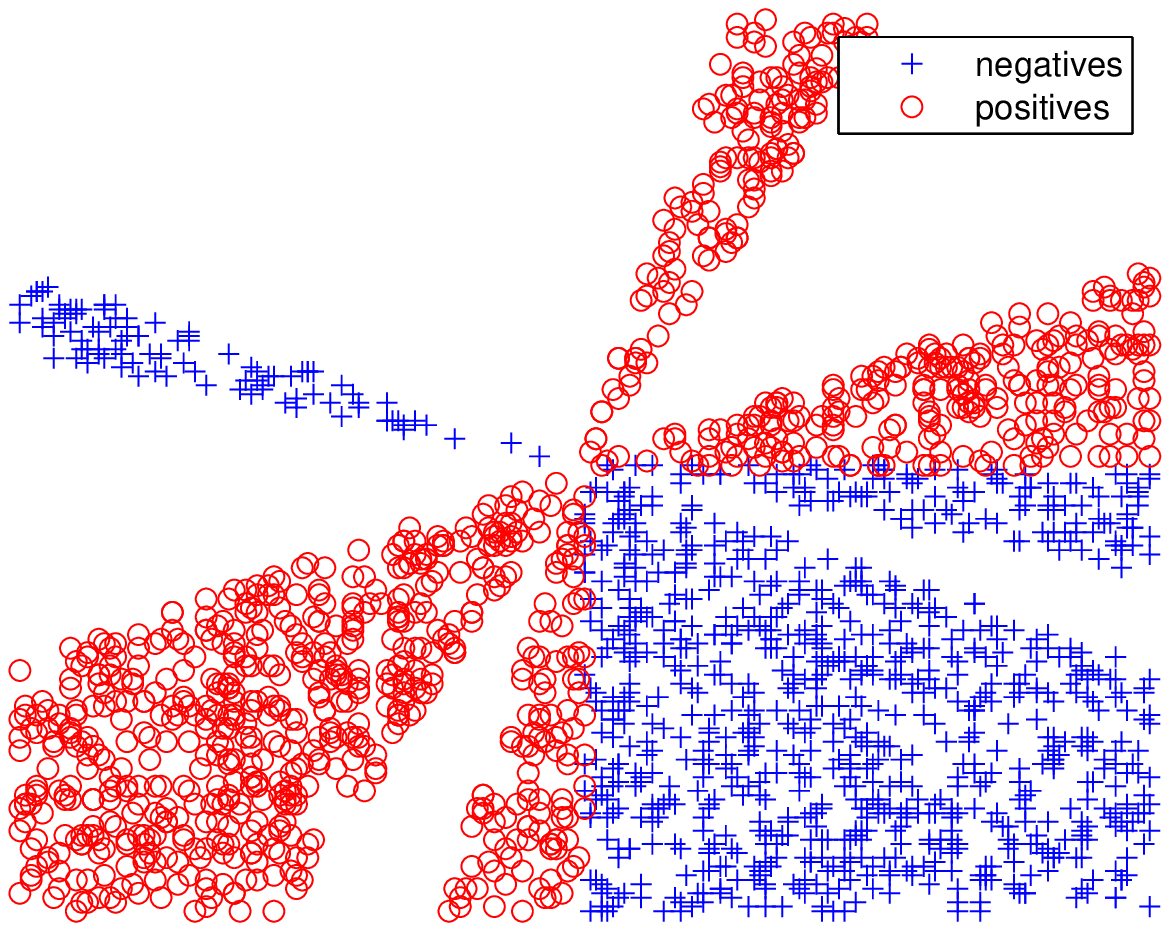,width=1.0in} &
\hspace{-8mm} \psfig{figure=\MYF/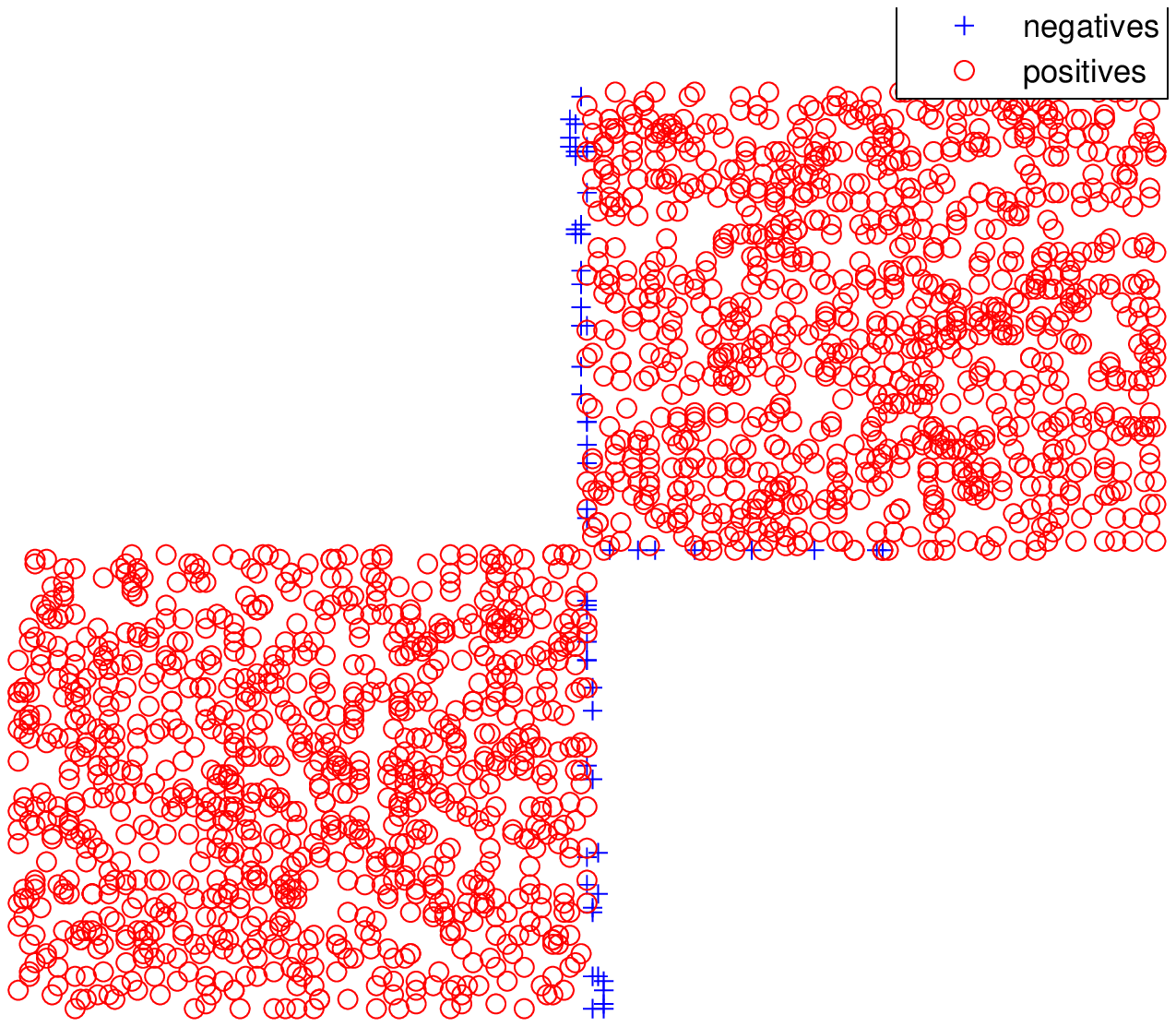,width=1.0in} \\
(a) & (b) & (c) & (d) \\
\end{tabular}
\end{center}
\vspace{-3mm}
\caption{The xor problem. (a) shows the positive and negative points.
(b) are the points classified as positives by the AdaBoost algorithm using
100 decision stump weak classifiers. (c) shows
points classified as positives by the Ada-Ada algorithm which will be
discussed later. (d) shows the positive points classified
by the Ada-Or classifier using 10 OrBoost weak classifiers.}
\label{fig:toy}
\end{figure}

Classification algorithms such as
decision tree~\cite{C4.5,CART}, neural networks~\cite{Bishop},
support vector machine (SVM) have been widely used in many areas.
The classification procedure in many of these algorithms can be understood as performing
reasoning using logic operators (and, or, not), with deterministic or probabilistic formulations.
The recent development of the AdaBoost algorithm~\cite{AdaBoost}
has particularly advanced the performance of many applications in the field.
We focus on the AdaBoost algorithm in this paper (also called
 boosting together with its variations~\cite{Breiman,Friedman98}).

Boosting algorithms have many advantages over the traditional classification algorithms.
Its asymptotical behavior when combining a large number of weak classifiers
is less prone to the overfitting problem.
Once trained, a boosting algorithm performs weighted sum on the
selected weak classifiers. This linear summation weakly performs
the `and' and `or' operations. In the discrete case, as long as the overall score
is above the threshold, a pattern is considered as positive. This may include
a combinotory combinations of the conditions. Some weak classifiers
may require to be satisfied together (`and'), and some may not as long a subset answer yes (`or').

In the literature, decision stump has been widely used as weak classifier 
due to its speed and small complexity.
However, decision stump does not have strong discrimination power. A comprehensive empirical study
for a wide variety of classifiers including SVM, Boosting (using decision-tree and decision-stump),
neural networks, and nearest neighboorhood, was reported in \cite{Caruana}.
Each decision stump corresponds to a thresholded feature ($\ge$ is changeable to $<$):
\begin{equation}
    h(F_j(x),tr)= \left\{ \begin{array}{ll}
      +1 & \textrm{if } F_j(x) \ge tr \\
      -1 & \textrm{otherwise} \\
      \end{array} \right.
\end{equation}
We call stump based AdaBoost algorithm {\em Ada-Stump} for the remainder of this paper.
Fig.~(\ref{fig:toy}.b) displays a failure example of the Ada-Stump.
We see that it can not deal with the `xor' patterns, even with 100 stumps. 

One solution to this problem is to adopt more powerful weak classifiers, such
as decision tree, to the boosting algorithm. It was proposed by several authors~\cite{Friedman98,Reyzin} and
we call it Ada-Tree here for notional convenience (it is different from the AdaTree method~\cite{AdaTree}).
However, using decision tree greatly increases the time and computational complexity of the boosting algorithm.
Many vision applications were trained on very large datasets with each sample having
thousands or even millions features~\cite{Viola}.
This limits the use of decision tree or CART, and Ada-Stump remains
mostly used in vision~\cite{Viola}.
In this paper, we show that Ada-Stump intrinsically can not deal with the `xor' problem.
We propose layered logic models for classification, namely {\em Ada-Or}, {\em Ada-And}, and {\em Ada-AndOr}.
The algorithm has several interesting properties:
(1) it naturally incorporates the `and', `or', and `not' relations in the algorithm; (2) it has much
more discrimination power than Ada-Stump; (3) it has much smaller computational complexity than tree
based AdaBoost with only slightly degraded classification performance.

A recent effort to combine `and' and `or' in AdaBoost has been proposed in~\cite{Dundar}. However, the
`and' and `or' relations are not naturally embedded in the algorithm and it requires very complex optimization
procedure in training. How the algorithm can be used for general tasks in machine learning
and computer vision is at best unclear.

We apply the proposed models, Ada-Or, Ada-And, and Ada-AndOr, on several typical datasets from the
Irvine repository and two challenging vision applications, object segmentation and pedestrian detection.
Among the models, Ada-AndOr performs the best nearly in all cases.
We observe significant improvements on all the datasets over Ada-Stump.
For pedestrian detection, the performance of Ada-AndOr is very close to HOG~\cite{HOG}
using simple Haar features, though the main objective of this paper
is not to develop a pedestrian detector.

\section{AdaBoost algorithm \label{sect:adaboost}}

In this section, we briefly review the AdaBoost algorithm and explain
why Ada-Stump fails on the `xor' problem.

\subsection{Algorithms and theory \label{sect:theory}}

Let $\{(x_i,y_i,D_1(i)), i=1...N\}$ be a set of training samples and $D_1(i)$ is the distribution
for each sample $x_i$. AdaBoost algorithm~\cite{AdaBoost} proposed by Freund and Schapire 
learns a strong classifier $H(x)=sign(\sum_{t=1}^T \alpha_t h_t(x))$, based on the training set,
by sequentially combining a number of weak classifiers.
We briefly give the general AdaBoost algorithm\cite{AdaBoost} below:

\begin{figure}[!htb]
\begin{center}
\begin{tabular}{|c|}
\hline
\begin{minipage}[c]{80mm}{
{\footnotesize
Given: $(x_1,y_1,D_1(1)),...,(x_N,y_N,D_1(N));\; y_i \in \{-1, 1 \}$

For $t=1,...,T:$
\begin{itemize}
  \item[$\bullet$] Train weak classifier using distribution $D_t$.
  \item[$\bullet$] Get weak hypothesis $h_t: \chi \to \{-1,+1 \}$.
  \item[$\bullet$] Calculate the error of $h_t: \epsilon_t=\sum_{i=1}^N D_t(i) \1_{(y_i \ne h_t(x_i))}$.
  \item[$\bullet$] Compute $\alpha_t=-\log \epsilon_t/(1-\epsilon_t)$.
  \item[$\bullet$] Update: $D_{t+1}(i) \leftarrow D_t(i) \cdot \exp (-\alpha_t y_i h_t(x_i))$ with $\sum_i D_{t+1}(i)=1$.
\end{itemize}
Output the the strong classifier: $H(x)=sign(\sum_{t=1}^T \alpha_t h_t(x))$.
}
}\end{minipage} \\
\hline
\end{tabular}
\end{center}
\caption{Discrete AdaBoost algorithm. $\1$ is an indicator function. The variations to the
AdaBoost algorithms such as arc-gv~\cite{Breiman}, RealBoost and GentalBoost~\cite{Friedman98}
differ mostly from the way $h_t$ and $\alpha_t$ are computed.}
\label{fig:AdaBoost}
\end{figure}

The AdaBoost algorithm minimizes the total error $\sum_i e^{-\sum_{t=1}^T \alpha_t y_i h_t(x_i)}$
by sequentially selecting $h_t$ and computing $\alpha_t$ in a greedy manner. At each step, it
is to minimize
\[
    E_t = \sum_i D_t(i) e^{- \alpha_t y_i h_t(x_i)}
\]
by coordinate descent:

(1) Select the best weak classifier from the candidate pool which minimizes $E_t$.

(2) Compute $\alpha_t$ by taking $\frac{d E_t} {d \alpha_t}=0$, which yields
\[
	 \alpha_t=-\log \epsilon_t/(1-\epsilon_t).
\]

A very important property of AdaBoost is that after a certain number
rounds, the test error still
goes down even the training error is not improving~\cite{BreimanArc}. This makes AdaBoost
less prone to the overfitting problem than many other classifiers.
Schapire et al.~\cite{Margin} explained
this behavior of AdaBoost from the margin theory perspective. For any data $(x,y)$,
\[
   margin(x,y) = \frac{y \sum_t \alpha_t h_t(x)} {\sum_t \alpha_t}.
\]
$margin(x,y)$ essentially gives the confidence of the estimation $y$ to $x$.
For any given $\theta$, the overall test error is bounded by
\begin{equation}
   error_{test} (x, y) \le p[margin(x,y) \le \theta] + \tilde{O}(\sqrt{ \frac{d}{m \theta^2}}),
\label{eq:margin}
\end{equation}
where $d$ is the VC dimension of the weak classifier and $m$ is the number of training
samples. Eqn.~(\ref{eq:margin}) shows three directions to reduce the test error:
(1) increase the margin (related to training error but not exactly the same);
(2) reduce the complexity of the weak classifier;
(3) increase the size of training data.

Moreover, it is shown~\cite{Friedman98} that AdaBoost and its variations are asymptotically approaching
the posterior distribution (there are still some debates about
this probabilistic formulation of the AdaBoost algorithm). 

\begin{equation}
	p(y|x) = \frac{e^{2 y \sum_t \alpha_t h_t(x)}}{1+ e^{2 y \sum_t \alpha_t h_t(x)}}.
\label{eq:logistic}
\end{equation}
The margin is directly tied to the discriminative probability.

\subsection{The xor problem}

It is well-known that the points shown in Fig.~(\ref{fig:toy}.a) as `xor' are not
linearly separable. The red and blue points are the positive
and negative samples respectively. Each weak classifier makes
a decision whether a point lies above or below a line passing the original.
Using this type of weak classifier, the AdaBoost algorithm is not able to
separate the red points from the blue ones. It is easy to verify.
For any positive sample $(x_1, x_2)$ with $H(x_1,x_2)=\sum_{t=1}^T \alpha_t h_t(x_1, x_2) > 0$,
then $(-x_1, -x_2)$ is a positive sample also. However
\[
h_t(-x_1, -x_2) = -h_t(x_1, x_2), \forall h_t
\]
and therefore $H(-x_1,-x_2)<0$.

\begin{figure}[!htbp]
\begin{center}
\begin{tabular}{c}
\psfig{figure=\MYF/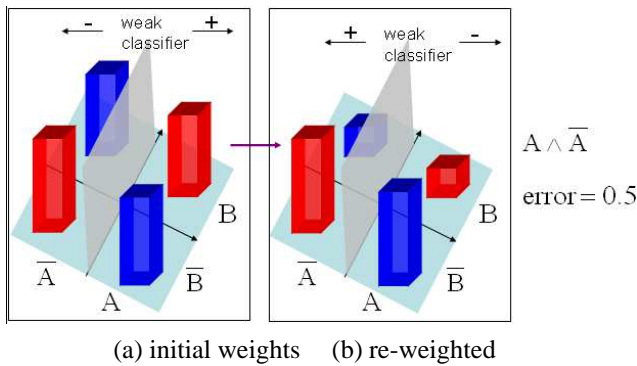,width=3.3in} \\
\end{tabular}
\begin{tabular}{cc}
(a) initial weights & (b) re-weighted \\
\end{tabular}

\end{center}
\caption{The re-weighting scheme in the AdaBoost to cause a deadlock.}
\label{fig:reweight}
\end{figure}

Let
\[
    A = \{(x_1,x_2), x_1>0 \} \; and \; \bar{A} = \{(x_1,x_2), x_1<0 \}, and
\]
\[
    B = \{(x_1,x_2), x_2>0 \} \; and \; \bar{B} = \{(x_1,x_2), x_2<0 \}.
\]
We denote $\land$, $\lor$, $\bar{}$ as the `and', `or', and `not' operations respectively.
Thus, the positive samples in Fig.~(\ref{fig:toy}.a) can be denoted by
\[
    (A \land B) \lor (\bar{A} \land \bar{B}), \; or \; (A \lor \bar{B}) \land (\bar{A} \lor B).
\]

One of the key properties in AdaBoost is that it re-weights
the training samples after each round by giving higher weights
to those which were not correctly classified by the previous
weak classifiers. We take a close look at the re-weighting scheme to
the points in Fig.~(\ref{fig:toy}.a).
Initially, all the samples
receive equal weights, shown in Fig.~(\ref{fig:reweight}.a).
For any weak classifier (line passing the origin), the error is $\epsilon=0.5$ which
means that they are equally bad.
In a computer simulation the value is usually slightly smaller than $0.5$ since the
training points are discretized samples. Once a weak classifier is selected, e.g.,
the line $x_1>0$ ($A$), then positive samples $(A \land B)$ and negative samples
$(\bar{A} \land B)$ are correctly classified, and they will receive lower weights.
Fig.~(\ref{fig:reweight}.b) shows the weights for the samples after the first step of the
AdaBoost. Clearly, the weak classifier to minimize the error for this round
would be $x_1<0$ ($\bar{A}$), which is a contradictory decision to the previous weak classifier ($A$).
The re-weighted points after this round essentially lead the situation back to Fig.~(\ref{fig:reweight}.a).
The combination of the two weak classifiers is $A \land \bar{A}=\phi$ where $\phi$ denotes
an empty set. The algorithm then keeps repeating the same procedure, which is a deadlock.
Due to this reason, AdaBoost is sensitive to outliers since
it keeps giving high weights to those miss-classified samples.

\subsection{Possible solutions}
The previous section shows that Ada-Stump cannot solve the `xor' problem (on the line features passing
the origin). The AdaBoost algorithm makes an overall decision based on a weighted sum
$H(x)=sign(\sum_{t=1}^T \alpha_t h_t(x))$.
It weakly performs the `and', and `or' operations on the weak classifiers. The `not' is often embedded in the stump classifier
by switching $>$ and $<$. We assume that all types of weak classifiers have the aspect of
`not' and we focus on `and', and `or' operations for the rest of this paper.

There are several possible ways to improve the algorithm:
\begin{enumerate}
\item Designing hyper features to allow the patterns to be linearly separable. For example, in the `xor'
case, it could be $x_1 \times x_2$. However, (1) it
is often very hard to find the meaningful features which will nicely separate the positive
and negative samples; (2) complex features often lead to the over-fitting problem.
\item Introducing the explicit `and' and `or' relations into the AdaBoost.
\end{enumerate}

We can put `and's on top of `or's, or vice versa, or completely mix the two together. The probabilistic
boosting tree (PBT) algorithm~\cite{PBT} is one way of recursively combining `and's with `or's.
The disadvantages of PBT however are: (1) it requires longer training time than cascade and, (2) it produces 
complex classifier and may lead to overfitting (like the decision tree).
Another solution is to build weak classifiers
with embedded `and' and `or' operations.
Using decision tree~\cite{C4.5} as weak classifiers has been described in several papers~\cite{Friedman98,Reyzin}.
However, each tree is a complex classifier and it requires much longer time
in training than the stump classifier. Also, it has more algorithm complexity than decision stump.

\section{Layered logic classifiers \label{sect:adaorboost}}

Eqn.~(\ref{eq:logistic}) shows that the AdaBoost algorithm is essentially approaching
a logistic probability by
\[
	p(y|x) \propto e^{y \sum_t \alpha_t h_t(x)} \propto \prod_t e^{y \alpha_t h_t(x)}.
\]
The overall discriminative probability is a product of the probability of each
$h_t$. Depending upon its weight $\alpha_t$, each $h_t$ makes a direct impact on $p(y|x)$.
Using decision tree requires much longer time than stump
classifier. This is particularly a problem in vision as we often face millions of image samples with
each sample having thousands features.

Instead of using one layer AdaBoost, we can think of using two-layer AdaBoost
with the weak classifier being stronger than decision stump, but simpler than decision tree.
One idea might be to use Ada-Stump as weak classifier for the AdaBoost again, which we
call Ada-Ada-Stump, or Ada-Ada for short notation. However, Ada-Ada still somewhat performs a linear summation and
has difficulty on the `xor' as well. Fig.~(\ref{fig:toy}.c) shows the positives classified by
Ada-Ada with 50 weak classifiers of Ada-Stump, which itself has 5 stump weak classifiers.
It is a failure example.
It is worth to mention that one can indeed to make Ada-Ada work on these points by using
very tricky strategies of randomly selecting a subset of points in training.
However this greatly increases the training complexity and the procedures are not general.

Our solution is to propose  AndBoost and OrBoost algorithms in which
the `and' and `or' operations are explicitly engaged.
We give detailed descriptions below.

\subsection{OrBoost \label{sect:orboostd}}
For a combined classifier, we can use the `or' operation directly by
\begin{equation}
   H(x) = sign (h_1(x) \lor ... \lor h_T(x)),
\end{equation}
where
\begin{equation}
    h_i(x) \lor h_j(x) = \left\{ \begin{array}{ll}
      +1 & if \; h_i(x)=+1 \; or \; h_j(x)=+1\\
      -1 & \textrm{otherwise} \\
      \end{array} \right.
\end{equation}

\begin{figure}[!htb]
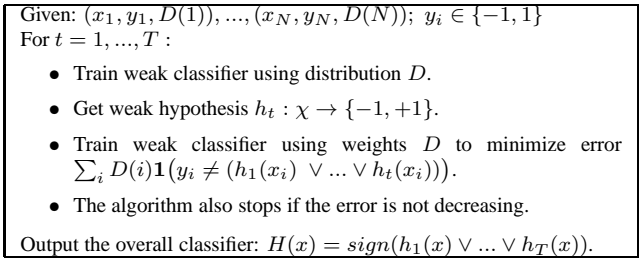

\begin{center}
\begin{tabular}{|c|}
\hline
\begin{minipage}[c]{80mm}{
{\footnotesize
Given: $(x_1,y_1,D(1)),...,(x_N,y_N,D(N));\; y_i \in \{-1, 1 \}$

For $t=1,...,T:$
\begin{itemize}
  \item[$\bullet$] Train weak classifier using distribution $D$.
  \item[$\bullet$] Get weak hypothesis $h_t: \chi \to \{-1,+1 \}$.
  \item[$\bullet$] Train weak classifier using weights $D$ to minimize error $\sum_i D(i) \1 \big (y_i \ne (h_1(x_i)\; \lor ... \lor h_t(x_i)) \big)$.
  \item[$\bullet$] The algorithm also stops if the error is not decreasing.
\end{itemize}
Output the overall classifier: $H(x) = sign (h_1(x) \lor ... \lor h_T(x))$.
}
}\end{minipage} \\
\hline
\end{tabular}
\end{center}
\caption{OrBoost algorithm.}
\label{fig:orboost_d}
\end{figure}

Fig~(\ref{fig:orboost_d}) gives the detailed procedure
of the OrBoost algorithm, which is straight forward to implement.
The overall classifier is a set of `or' operations on weak classifier, e.g.
decision stump, and it favors positive answer. If any weak classifier provides
a positive answer, then the final decision is
positive, regardless of what other weak classifier will say.
Unlike in the AdaBoost algorithm where mis-classified samples are given
higher weights in the next round, OrBoost gives up some samples quickly
and focus on those which can be classified correctly.
This helps to solve the deadlock situation in AdaBoost
shown in Fig.~(\ref{fig:reweight}).

\begin{figure}[!htbp]
\begin{center}
\begin{tabular}{c}
\psfig{figure=\MYF/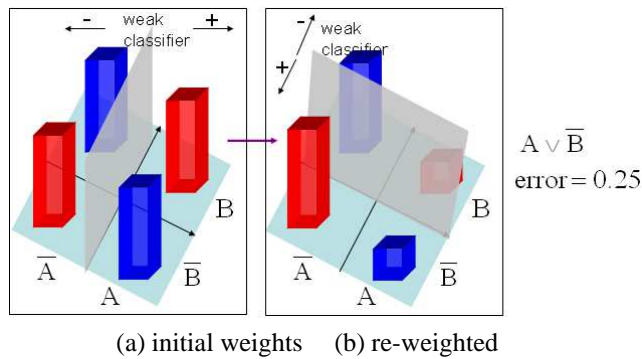,width=3.3in} \\
\end{tabular}
\begin{tabular}{cc}
(a) initial weights & (b) re-weighted \\
\end{tabular}

\end{center}
\caption{The re-weighting scheme in the OrBoost algorithm which breaks the deadlock
in the AdaBoost algorithm.}
\label{fig:reweight_or}
\end{figure}

Fig.~(\ref{fig:reweight_or}) shows the feature selection and re-weighting steps by the OrBoost algorithm
for the xor problem. The first weak classifier is selected the same as before ($x_1 > 0 = A$).
However, positives $AB$ and negatives $\bar{A}B$ receive low weights in the AdaBoost
since they have been classified correctly. This creates a deadlock.
In OrBoost, the situation is different. Note that although the weights for all the
samples $D$ are fixed, the error evaluation function $\sum_i D(i) \1 \big (y_i \ne (h_1(x_i)\; \lor ... \lor h_t(x_i)) \big)$
affects how $D(i)$ plays a role. This is similar to the re-weighting scheme in the AdaBoost.
For example, positives $AB$ and negatives $A\bar{B}$ have been
classified as positives by the first weak classifier, $x_1 > 0 = A$. The errors on $AB$ and $A\bar{B}$
are therefore decided already regardless what the later weak classifiers will be.
Therefore, the second weak classifier would be $x_2 < 0 = \bar{B}$. The total error
by the two combined weak classifiers is $0.25$. 

\subsection{AndBoost \label{sect:andboost}}
If we swap the labels of the positives and negatives in training,
the `or' operations in OrBoost can be directly turned into `and' operations since
\[
   A \land B = \bar{A} \lor \bar{B}.
\]
However, for a given set of the training samples, `and' operations may
provide complementary decisions to the `or' operations.
Similarly, we can use the `and' operation directly by
\begin{equation}
   H(x) = sign (h_1(x) \land ... \land h_T(x)),
\end{equation}
where
\begin{equation}
    h_i(x) \land h_j(x) = \left\{ \begin{array}{ll}
      +1 & if \; h_i(x)=+1 \; and \; h_j(x)=+1\\
      -1 & \textrm{otherwise} \\
      \end{array} \right.
\end{equation}

Therefore, we can design an AndBoost algorithm in Fig.~\ref{fig:andboost} which
is very similar to the OrBoost algorithm in Fig.~\ref{fig:orboost_d}.
\begin{figure}[!htb]
\begin{center}
\begin{tabular}{|c|}
\hline
\begin{minipage}[c]{80mm}{
{\footnotesize
Given: $(x_1,y_1,D(1)),...,(x_N,y_N,D(N));\; y_i \in \{-1, 1 \}$

For $t=1,...,T:$
\begin{itemize}
  \item[$\bullet$] Train weak classifier using distribution $D$.
  \item[$\bullet$] Get weak hypothesis $h_t: \chi \to \{-1,+1 \}$.
  \item[$\bullet$] Train weak classifier using weights $D$ to minimize error $\sum_i D(i) \1 \big (y_i \ne (h_1(x_i)\; \land ... \land h_t(x_i)) \big)$.
  \item[$\bullet$] The algorithm also stops if the error is not decreasing.
\end{itemize}
Output the overall classifier: $H(x) = sign (h_1(x) \land ... \land h_T(x))$.
}
}\end{minipage} \\
\hline
\end{tabular}
\end{center}
\caption{AndBoost algorithm.}
\label{fig:andboost}
\end{figure}

The performance of the AndBoost on the `xor' problem is the same as the OrBoost algorithm.

\subsection{AdaOrBoost}

\begin{figure*}[!htbp]
\begin{center}
\begin{tabular}{c}
\psfig{figure=\MYF/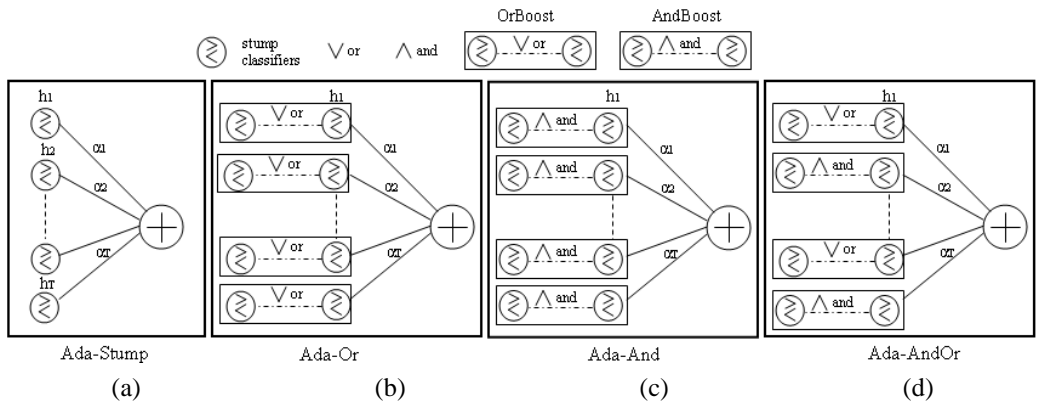,width=5.4in} \\
\end{tabular}
\begin{tabular}{cccc}
(a) \hspace{26mm} & (b) \hspace{26mm} & (c) \hspace{26mm} & (d)\\
\end{tabular}
\end{center}
\vspace{-3mm}
\caption{Two-layer logic models. (a) shows a traditional Ada-Stump algorithm. (b) displays
the Ada-Or algorithm with OrBoost being the choice of weak classifier in the second layer.
(c) and (d) give the illustrations of Ada-And and Ada-AndOr respectively.}
\label{fig:andor}
\end{figure*}

After the introduction of the OrBoost and AndBoost algorithms, we
are ready to discuss the proposed layered models. We simply use a two-layer
AdaBoost algorithm with the weak classifiers in the second layer being the choice of OrBoost, AndBoost, or
both. We call the models, {\em Ada-Or}, {\em Ada-And}, and {\em Ada-AndOr} respectively.

There are two levels of weak classifiers now. For Ada-Or, the OrBoost is its weak classifier.
For OrBoost, any type of classifier can be its weak classifier.
To keep the complexity of OrBoost and AndBoost under check, we simply use the decision stump.
As we mentioned before the `not' operation is naturally embedded in
the decision stump.
Therefore, the Ada-AndOr has all the aspects of logic operations, `and', `or', and `not'.
Again, we call the weak classifiers in OrBoost and AndBoost {\em operations to} avoid confusion.

Fig.~(\ref{fig:toy}.b) shows the points which are classified by Ada-Stump using
100 stump weak classifiers. This failure example verifies our earlier claim for the `xor' pattern.
Fig.~(\ref{fig:toy}.d) shows the result by Ada-Or using 10 OrBoost weak classifiers, in which
there are 2 or operations. As we can see, the positive samples
have been classified correctly.
Fig.~(\ref{fig:toy}.c) gives the result by Ada-Ada.

The margin theory of Ada-Or, Ada-And, and Ada-AndOr still follows the same as pointed by Schapire et al.~\cite{Margin}
in eqn.~(\ref{eq:margin}). The complexity $d$ of weak classifier is decided by the OrBoost and AndBoost algorithms,
which are just a sequence of `or' operations or `and' operations. It is slightly more complex than decision stump,
but much simpler than decision tree or CART.
It is worth to mention that both OrBoost and AndBoost include a special case where only one operation presents.
This happens when the training error is not improving by adding the second operation. Therefore, stump
classifier is also included in OrBoost and AndBoost, if stump is the choice of operator.

\subsection{Experiments \label{sec:sampling}}

There are several major issues we are concerned with for the choice of different
classifiers for applications in machine learning and computer vision.

\begin{enumerate}
\item {\em Classification power}: This is often referred to as
training error or margin in eqn.~(\ref{eq:margin}).
A desirable classifier should produce low error and large margin on the training data.
\item {\em Low complexity}: This is often called VC dimension~\cite{SVM} and a classifier
with small VC dimension often has a good generalization power, small difference between the training error
and test error.
\item {\em Size of training data}: In the VC dimension and margin theory, the overall
test error is also greatly decided by the availability of training data. The more training data
we have and the classifier can handle, the smaller difference is between training error and
test error. In reality, we often do not have enough training data since collecting them
is not a easy task. Also, some none-parametric classifiers can only deal with limited amount
of training data since they work on the kernel space, which explodes on large size data.
\item {\em Efficient training time}: For many applications in computer vision and data mining,
the training data size can be immense and each data sample also has a large number of features.
This demands an efficient classifier in training also. Fast training is more required in online learning
algorithms~\cite{Oza} which has recently received many attentions in tracking.
\item {\em Efficient test time}: Judging the performance of a classifier is ultimately done in the
test stage. A classifier is expected to be able to quickly give an answer. For many modern
classifiers, this is not particularly a problem.
\end{enumerate}
The first three criterion collectively decide the test error of a classifier.
Another major factor affecting the performance a classifier is feature design.
If the intrinsic features can be found, different types of classifiers will
probably have a similar performance. However, the discussion of feature design is out of the scope of
this paper. Next, we focus on the performance of AdaOrBoost with comparison to the
other classifiers.

\subsection{Results on UCI repository datasets}

One of the reasons that the AdaBoost algorithm is widely used is 
due to nice generalization power. Schapire et al. gave an explanation based on the
margin theory after Breiman~\cite{BreimanArc} observed an interesting behavior of AdaBoost:
the test error of AdaBoost further asymptotically goes down even the training error is not decreasing.
This was explained
in the margin theory as to increase the margin with more weak classifiers combined.
Brieman~\cite{Breiman} then designed an algorithm called `arc-gv' which
tries to directly maximize the minimum margin in
computing the $\alpha_t$ for AdaBoost. The experimental results were however contradictory to the
theory since arc-gv produces bigger test error than AdaBoost. Reyzin and Schapire~\cite{Reyzin} tried to
explain this finding and showed that the bigger test error by arc-gv was
indeed due to the use of complex weak classifier, CART. Next we compare
Ada-Or, Ada-And, and Ada-AndOr with arc-gv and AdaBoost using CART and decision stump.

We use the same datasets shown in Reyzin and Schapire~\cite{Reyzin}, which are
all from the UCI repository: breast cancer, ionosphere, ocr49 and splice.
The datasets have been slightly modified the same way as in~\cite{Reyzin}.
The two splice categories were merged into one in the splice dataset to create
two-class data. Only digits 4 and 9 from the NIST database were used in the ocr49 dataset.
The cancer, ion, ocr49 and splice then have 699, 351, 6000, 3175 data points respectively.
Each sample usually has $20-60$ features, depending upon what dataset it belongs to.
The data samples are randomly split into training and testing for 10 trials. Table~\ref{tb:uci}
shows the corresponding numbers.
\begin{table}[!htb]
\begin{center}
{\scriptsize
\begin{tabular}{|c|c|c|c|c|}
\hline
 & cancer & ion & ocr 49 & splice \\
 \hline
training & 630 & 315 & 1000 & 1000 \\
 \hline
test & 69 & 36 & 5000 & 2175 \\
\hline
 \end{tabular}
}
\end{center}
\caption{The sizes of training and test data of four datasets from UCI repository.
The training and test data samples are randomly selected.}
\label{tb:uci}
\end{table}

\begin{figure}[!htbp]
\begin{center}
\begin{tabular}{cc}
\hspace{-6mm} \psfig{figure=\MYF/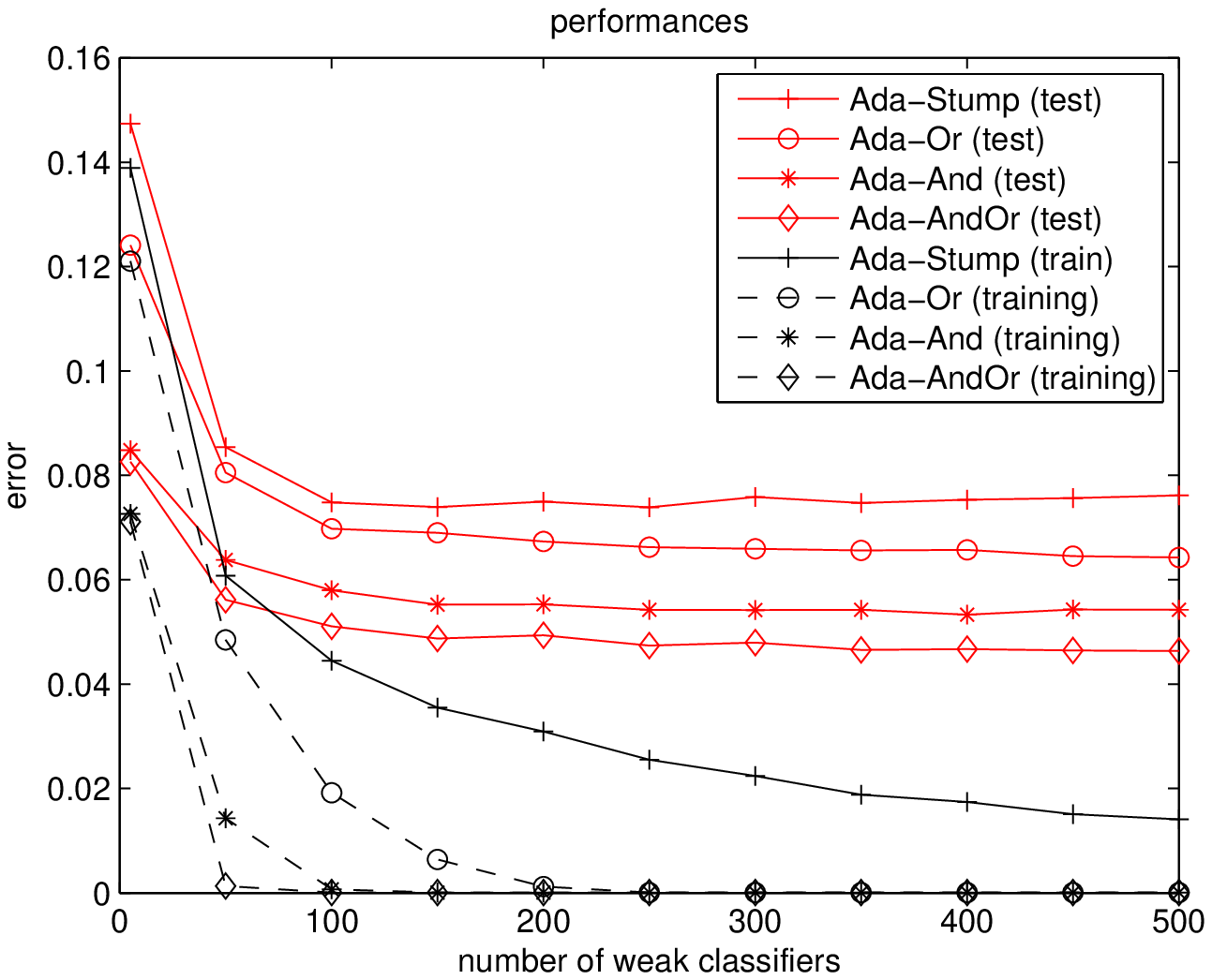,width=1.8in} & 
\hspace{-6mm} \psfig{figure=\MYF/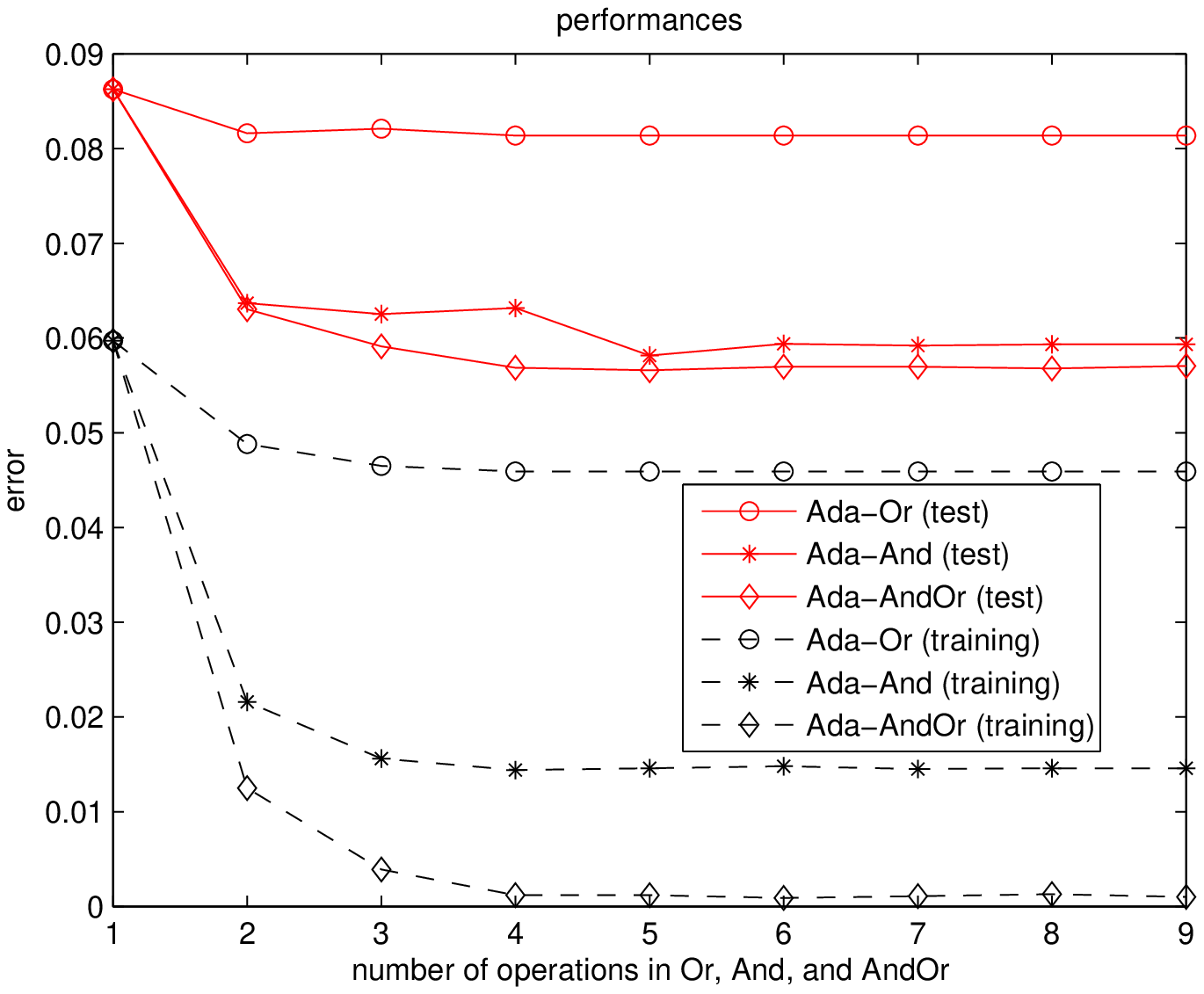,width=1.8in} \\
(a) & (b) \\
\end{tabular}
\end{center}
\caption{Training and test errors on the splice dataset by proposed models using different number
of weak classifiers and operations. (a) displays comparison with different number of weak classifiers.
Each curve is is averaged over 10 trials by randomly 
splitting the dataset into training and test samples. 
(b) shows comparison of using different operations. Each algorithm
uses 50 weak classifiers.}
\label{fig:different_op}
\end{figure}

To illustrate the effectiveness of the layered models, we first
compare its results to those by Ada-Stump.
Though there are other alternatives such as RealBoost and GentalBoost~\cite{Friedman98},
decision stump remains being widely adopted in the AdaBoost implementation.
Fig.~(\ref{fig:different_op}.a) shows the training and test errors on the splice dataset
by Ada-Stump, Ada-Or, Ada-And, and Ada-AndOr using different number of weak classifiers.
In the implementation of OrBoost and AndBoost, we use 5 `or' operations.
Each curve is averaged over 10 trials by randomly selecting 
1000 samples for training and 2175 samples for testing.
The Ada-AndOr gives the best performance among all.
We also observe that the differences between the training and test errors
for Ada-Stump and others are very similar.
The results for real-world vision applications also show similar behavior of Ada-Or, Ada-and, and Ada-AndOr.
This suggests that the OrBoost and AndBoost algorithms are having similar
generalization power as decision stump.

To show how the use of different number of operations is affecting the performance,
we conduct another experiment on the splice dataset.
We plot out the training and test errors by using 50 weak classifiers with varying number of operations.
The overall performance of the models, both in training and testing, is not
improving too much with more than 3 operations shown in Fig.~(\ref{fig:different_op}.b). Similar observations apply to
other datasets as well. This suggests that the significant improvement
can be achieved without introducing too much overhead.

\begin{table*}[!htb]
\begin{center}
{\scriptsize
\begin{tabular}{|c|c|c|c|c|c|c|}
\hline
 & $\frac{arc-gv-CART}{arc-gv-stump}$ & $\frac{Ada-CART}{Ada-Stump}$ & $\frac{Ada-Stump (2500)}{Ada-Stump}$  & $\frac{Ada-Or}{Ada-Stump}$ & $\frac{Ada-And}{Ada-Stump}$ & $\frac{Ada-AndOr}{Ada-stump}$ \\
 \hline
breast cancer & $73.3\%$ & $57.3\%$ & $80.7\%$ & $57.4\%$ & $48.2\%$ & $56.0\%$ \\
\hline
ionosphere & $74.7\%$ & $36.1\%$ & $136.5\%$ & $63.6\%$ & $87.8\%$ & $66.7\%$\\
\hline
ocr 49 & $37.3\%$ & $32.5\%$ & $91.4\%$ & $57.2\%$ & $54.8\%$ & $38.2\%$\\
\hline
splice & $47.8\%$ & $46.8\%$ & $113.6\%$ & $83.8\%$ & $68.4\%$ & $60.8\%$ \\
\hline
 \end{tabular}
}
\end{center}
\caption{Test error ratios on the UCI datasets by arc-gv-CART, Ada-CART, Ada-Or, Ada-And, and
Ada-AndOr over Ada-Stump. 500 weak classifiers are used in all cases except for Ada-Stump (2500)
and Ada-Stump, which use 2500 and 100 stumps respectively.
Ada-Or, Ada-And, and Ada-AndOr all contain 5 operations in the OrBoost and AndBoost which have roughly 2500
stumps for each. Ada-AndOr significantly outperforms Ada-Stump, and it shows to be comparable to
arg-gv-CART, and is only a bit worse than Ada-CART.
}
\label{tb:comparison}
\end{table*}


It has been suggested~\cite{Friedman98,Breiman,Reyzin} that the best performance of boosting algorithm
is achieved by AdaBoost using decision tree~\cite{C4.5} or CART~\cite{CART}.
Some of the confusions about generalization (test) error based on the margin theory has
recently been clarified by Reyzin and Schapire~\cite{Reyzin}. In table~(\ref{tb:comparison}),
we compare the our algorithms with AdaBoost and arc-gv using decision tree.
For a fair comparison, we show the improvement of AdaOrBoost, arc-gv using CART, and AdaBoost
using CART over those using decision stump.
Table~(\ref{tb:comparison}) shows the error ratio. As we can see, the improvement of AdaOrBoost
is comparable to arc-gv using CART, but is worse than Ada-CART.
However, each CART, after tree pruning, has around 16 leaf nodes with the tree depth being around 7.
Therefore, the complexity of CART is much bigger than that of OrBoost and AndBoost.
This is particularly an issue for applications in vision as the training data is massive with each
data sample having thousands or even millions of features. The good performance of 
Ada-CART is achieved using an average of $7$ levels of tree.
This greatly limits its usage in many vision applications and leaves
the decision stump classifiers still being currently widely used~\cite{Viola}.

\begin{figure}[!htbp]
\begin{center}
\begin{tabular}{c}
\psfig{figure=\MYF/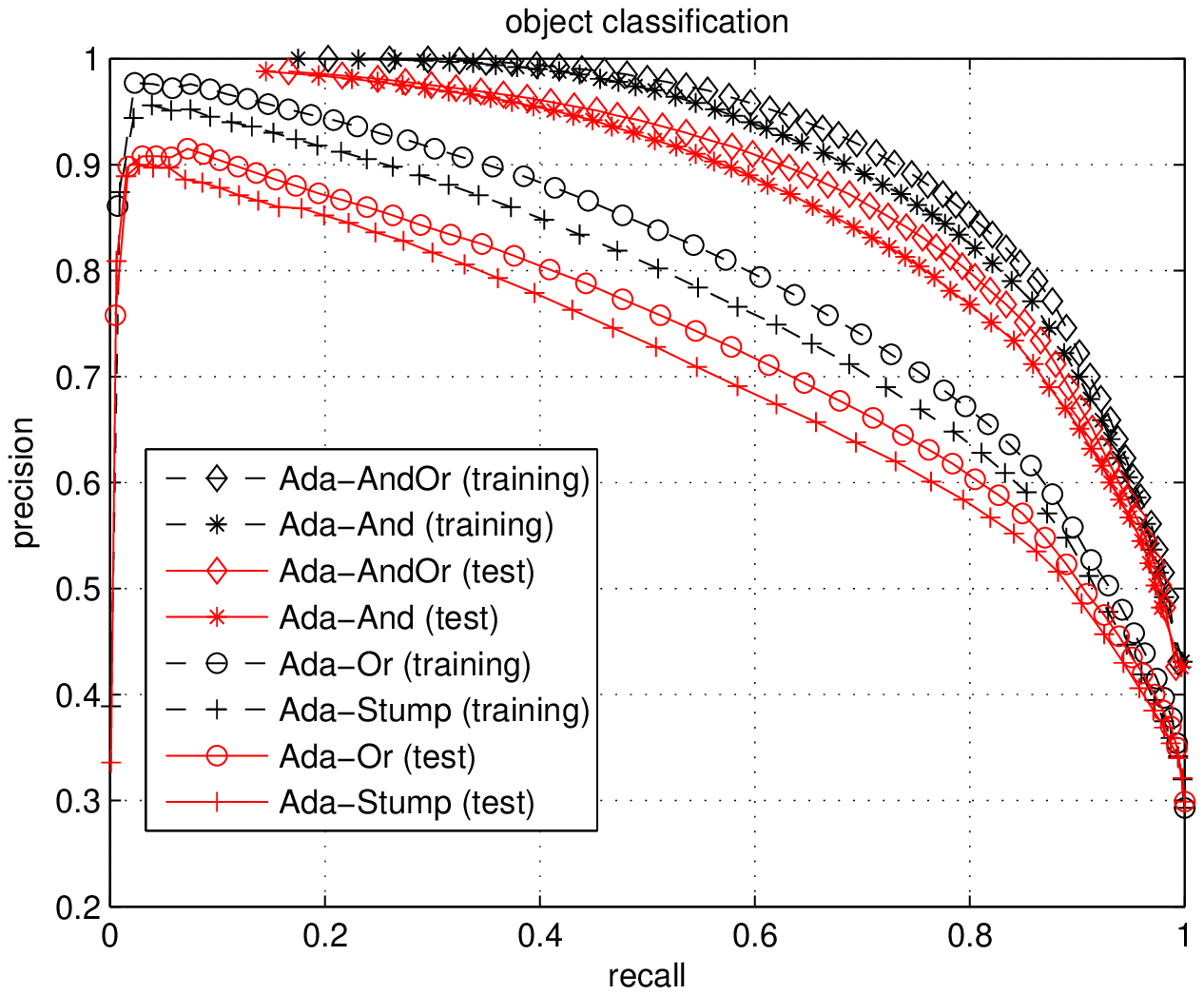,width=3.0in} \\
(a) \\
\psfig{figure=\MYF/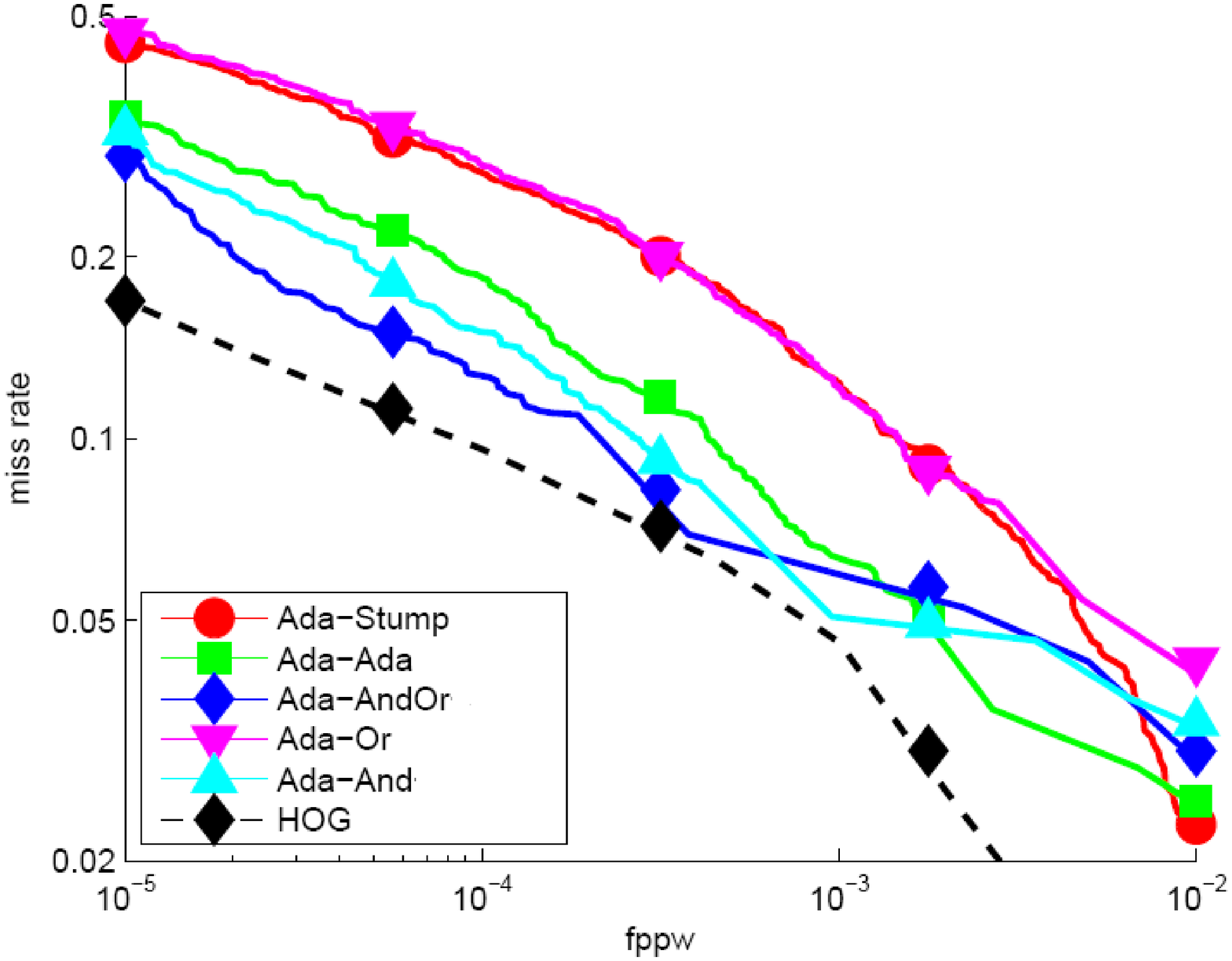,width=2.4in} \\
(b) \\
\end{tabular}
\end{center}
\vspace{-3mm}
\caption{Precision and recall curves for the horse segmentation and error
curve for pedestrian detection. (a) Shows the training and test errors by
decision stump based Ada-Stump, Ada-Or, Ada-And, and Ada-AndOr. (b) displays
the error curve by various classifiers. 3 operations are used for
all classifiers. Ada-AndOr achieves the
best result in both the cases among all Ada-. The horse segmentation
result (F=0.8) outperforms that reported in~\cite{Ren} (F=0.66) and
the pedestrian detection result is slightly worse than HOG~\cite{HOG}.
}
\vspace{-5mm}
\label{fig:detection}
\end{figure}

To illustrate the effectiveness of proposed algorithms, we further demonstrate them in 
two challenging vision problems, object segmentation and pedestrian detection.

First, we demonstrate it on the Weizmann horse dataset~\cite{Borenstein}.
We use 328 images and use 126 for training and 214 for testing.
Each input image comes with a label map in which the
pixels on the horse body and background are labeled as $1$ and $0$ respectively.
Given a test image, our task is to classify all the pixels into horse or background.
In training, we take image patch of size $21 \times 21$ centered on every pixel
as training samples. The background and horse body
image patches are the negatives and positives respectively.
For each image patch, we compute around $10,000$ features such as the mean, variance,
and Haar responses of the original as well as Gabor filtered.
We implement a cascade approach~\cite{Viola} and implement several versions. One uses Ada-Stump and
others use Ada-Or and Ada-AndOr.
Each cascade node selects and fuses 100 weak classifiers.
All the algorithms use an identical set of features and bootstrapping procedure.
Fig.~(\ref{fig:detection}.a) shows the precision and recall curve of the algorithms
on the training and test images. We observe similar result as that for the UCI repository
datasets. Ada-AndOr improves the results over Ada-Stump by a considerable amount.
The differences between the training and test errors are nearly the same in this cascade setting as well.
The F-value of the results by Ada-AndOr is around 0.8 which is better than the number 0.66 reported
in~\cite{Ren} which uses low and middle level information.

Next, we show the Ada-AndOr algorithm for pedestrian detection on dataset reported in~\cite{HOG}.
We use 8 level of cascade with different choices of weak classifiers for AdaBoost.
Fig.~(\ref{fig:detection}b) shows the results by Ada-Stump, Ada-Ada, Ada-Or, Ada-And, and Ada-AndOr.
The conclusion is nearly the same as before. Ada-AndOr achieves the best result among all with
Ada-And being on the second place. 
Though we are not specifically addressing the pedestrian detection problem here, the result is nevertheless
close to that by the well-known HOG pedestrian detector~\cite{HOG}.
However, we only use a set of generic Haar features without tuning the system specifically
for the pedestrian detection task.


\subsection{Conclusions}

Many of the classification problems in machine learning and computer vision
can be understood as performing logic operations combining `and', `or', and `not'.
In this paper, we have introduced layered logic classifiers. 
We show that AdaBoost can not solve the `xor' problem using decision stump type of weak classifiers.
We propose an OrBoost and AndBoost algorithms to study the `or' and `and' operations
respectively.
We demonstrate that the combined algorithm of two layers, Ada-AndOr, greatly outperformed
Ada-Stump which is widely used in the literature.
The improvement is significant in most the cases. We demonstrate the effectiveness of Ada-AndOr
on traditional machine learning datasets, as well as challenging vision applications.
Though decision tree based AdaBoost algorithm is shown to produce smaller
test error, its complexity in training often limits its usage.
The OrBoost and AndBoost algorithm only increases the time complexity slightly than
decision stump, but they significantly reduce the test error.
The Ada-AndOr algorithm is useful for a wide variety of applications
in machine learning and computer vision.

{\bf Acknowledgment}
This work is supported by NSF IIS-1216528 (IIS-1360566) and NSF CAREER award IIS-0844566 (IIS-1360568).

\bibliographystyle{ieee}

\begin{thebibliography}{1}

\bibitem{Bishop} C. M. Bishop, ``Neural networks for pattern recognition'', {\em Oxford University Press}, 1995.

\bibitem{CART} L. Breiman, J.H. Friedman, R.A. Olshen, C.J. Stone CJ,
``Classification and Regression Trees'', Chapman and Hall (Wadsworth, Inc.): New York, 1984.

\bibitem{Borenstein} E. Borenstein, E. Sharon and S. Ullman,
``Combining top-down and bottom-up segmentation'',
{\em Proc. IEEE workshop on Perc. Org. in Com. Vis.}, June 2004

\bibitem{BreimanArc} L. Breiman, ``Arcing classifiers'',
{The Annals of Statistics}, 26, pp 801-849, 1998.

\bibitem{Breiman} L. Breiman, ``Prediction games and arcing classifiers'',
{\em Neural Computation} 11, 1493-1517, 1999.

\bibitem{Caruana} R. Caruana and A. Niculescu-Mizil, ``An Empirical Comparison of Supervised Learning Algorithms'', {\em Proc. of ICML} , 2006.

\bibitem{HOG} N. Dalal and B. Triggs,
``Histograms of oriented gradients for human detection,''
{\em CVPR}, 2005.

\bibitem{Dundar} M. Dundar and J. Bi,
``Joint optimization of cascaded classifiers for computer aided detection'',
{\em Proc. of CVPR}, 2007.

\bibitem{BEL} P. Doll\'ar, Z. Tu, and S. Belongie,
``Supervised learning of edges and object boundaries'',
{\em Proc. of CVPR}, 2006.

\bibitem{Duda} R. O. Duda and P. E. Hart,
``Pattern classification'', Wiley Interscience, 2000.


\bibitem{Friedman98}  J. Friedman, T. Hastie and R. Tibshirani,
 ``Additive logistic regression: a statistical view of boosting'',
Dept. of Stat., Stanford  U. Te. Rep. 1998.

\bibitem{AdaBoost} Y. Freund and R. E. Schapire, ``A Decision-theoretic Generalization of
On-line Learning And An Application to Boosting'', {\em J. of Comp. and Sys. Sci.}, 55(1), 1997.

\bibitem{AdaTree} E. Grossmann, ``AdaTree: Boosting a Weak Classifier into a Decision Tree'',
{\em Proc. CVPR workshop on learning in computer vision and pattern recognition}, 2004.

\bibitem{Benchmark} D. Martin, C. Fowlkes, and J. Malik,
``Learning to detect natural image boundaries using local brightness, color and texture cues'',
{\em IEEE PAMI}, 26(5), 530-549, May 2004. 

\bibitem{Oza} N. Oza and S. Russell, ``Online Bagging and Boosting'',
{\em Proc. of 8th International Workshop on Artificial Intelligence and
Statistics}, 2001.

\bibitem{C4.5} J.R. Quinlan, ``Improved use of continuous attributes in C4.5'',
{\em J. of Art. Intell. Res.}, 4, pp. 77-90, 1996.

\bibitem{Ren} X. Ren, C. Fowlkes, and J. Malik, ``Cue integration in figure/ground labeling'',
{\em Proc. of NIPS}, 2005.

\bibitem{Reyzin} L. Reyzin and R. E. Schapire,
``How boosting the margin can also boost classifier complexity'',
{\em Proc. of the 23rd International Conference on Machine Learning}, 2006. 

\bibitem{Margin} R. E. Schapire, R. E. Freund, P. Bartlett, and W. S. Lee, 
``Boosting the margin: A new explanation for the effectiveness of voting methods. The Annals of
Statistics'', 26, pp. 1651-1686, 1998.

\bibitem{PBT} Z. Tu, ``Probabilistic boosting tree: Learning discriminative models
for classification, recognition, and clustering'', {\em Proc. of ICCV}, 2005.

\bibitem{SVM} V. Vapnik, ``Statistical Learning Theory''. Wiley-Interscience, 1998.

\bibitem{Viola} P. Viola and M. Jones, ``Robust Real-Time Face Detection'',
{\it Int'l J. of Comp. Vis.}, vol. 57, no. 2, pp. 137-154, 2004.

\end{thebibliography}

\end{document}